\pdfoutput=1
\pdftrailerid{}
\documentclass[10pt,twocolumn,letterpaper]{article}

\usepackage{cvpr}
\usepackage{graphicx}
\usepackage{microtype}
\usepackage{placeins}
\usepackage{tikz}
\usetikzlibrary{positioning,arrows.meta,calc}
\usepackage{multirow}

\definecolor{cvprblue}{rgb}{0.21,0.49,0.74}
\definecolor{paperaccent}{HTML}{0057E7}
\usepackage[
  breaklinks,
  colorlinks,
  pdftitle={},
  pdfauthor={},
  pdfsubject={},
  pdfkeywords={},
  pdfcreator={},
  pdfproducer={},
  linkcolor=paperaccent,
  citecolor=paperaccent,
  urlcolor=paperaccent
]{hyperref}

\def\paperID{}
\def\confName{CVPPA@ECCV}
\def\confYear{2026}

\newcommand{\etf}{\mathrm{ETF}}

\begin{document}
\raggedbottom

\title{Where Is the Bee? Detecting Tiny Pollinators\\
with a Single Collaborative-Head Transformer}

\author{Junsu Kim\qquad Seungryul Baek\\
UNIST, South Korea
}
\maketitle

\begin{abstract}
The CVPPA@ECCV~2026 BuzzSpot Challenge asks us to detect bees, bumblebees, hoverflies, and moths in $1920{\times}1080$ field keyframes. Its annotations carry 2 difficulties: the median box occupies $0.16\%$ of a frame, and bees account for $80\%$ of the labels. To cope with the small boxes, we compare 10 recorded detector configurations on held-out keyframes; plain Co-DINO with a Swin-L backbone has the highest mAP in this comparison, so we select it. Training then addresses the bee dominance in 2 ways: fine-tuning on a crop-mosaic pool in which the combined annotation share of the 3 rare classes rises from $19.9\%$ to $55.1\%$, and a class-weighted simplex equiangular tight frame (ETF) loss that pulls the projected states of matched decoder queries toward fixed class directions. The full schedule spans 12+3+2 epochs. Without inference-time ensembling or test-time augmentation, we rank first on FinalTest at $0.5062$ mAP@$[.5{:}.95]$. Code: {\normalfont\href{https://github.com/jjunsss/BeeHunters}{github.com/jjunsss/BeeHunters}}.
\end{abstract}

\section{Introduction}
\label{sec:intro}

The CVPPA@ECCV 2026 BuzzSpot Challenge asks us to localize bees, bumblebees, hoverflies, and moths in $1920{\times}1080$ field keyframes~\cite{BuzzSpot2026Challenge}. Its annotations make this hard in 2 ways: the median labeled box covers $0.16\%$ of a frame, the area of a $58{\times}58$-pixel square, leaving few pixels to localize an insect and identify its class; and bees contribute $9{,}611$ of the $12{,}000$ boxes, so most of the training signal comes from one class.

For the small boxes, Fig.~\ref{fig:teaser} compares plain Co-DINO against 9 YOLO and RTMDet configurations on held-out keyframes. Their training and inference protocols differ, so we use the comparison only to select the detector family; Co-DINO with Swin-L has the highest recorded mAP, and we use it in the final system~\cite{Zong2023CoDETR,Liu2021Swin}.

\begin{figure}[t!]
\centering
\includegraphics[width=0.98\linewidth]{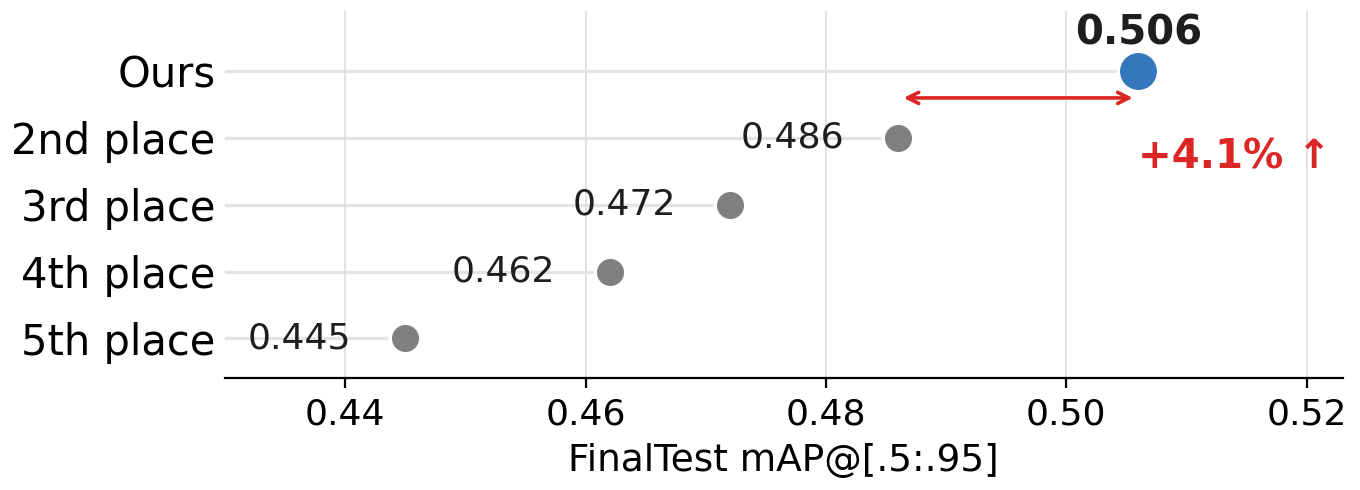}
\caption{\textbf{Official FinalTest leaderboard.}}
\label{fig:leaderboard}
\end{figure}

With the detector fixed, the bee dominance remains. We treat class exposure as a design variable~\cite{Kim2024ClassWiseBuffer}: object-centered crops of bumblebees, hoverflies, or moths receive higher sampling priority, and Mosaic augmentation~\cite{Bochkovskiy2020YOLOv4} tiles 4 of these sampled views into one training image, so the detector sees rare-class objects more often without new labels. For feature geometry, we transfer fixed-prototype alignment~\cite{Yang2022InducingNC,Yang2023NCFSCIL,Kim2025BeyondSyntheticReplays} to a class-weighted simplex equiangular tight frame (ETF) loss, which supervises each projected, Hungarian-matched decoder state toward a fixed direction for its class.

We organize these choices in a 3-stage schedule: 12 epochs adapting the detector on all labeled keyframes, 3 with class-aware crop Mosaic to shift class exposure, and 2 back on the original keyframes at a low learning rate, with the ETF loss active throughout. Without inference-time ensembling or test-time augmentation, a choice made for practical use, our checkpoint tops the official FinalTest leaderboard\footnote{\url{https://www.codabench.org/competitions/16441/}} in Fig.~\ref{fig:leaderboard} at $0.5062$ mAP@$[.5{:}.95]$.

\section{Method}
\label{sec:method}

We select the detector on held-out mAP (Sec.~\ref{sec:codino}), tailor its matched-query supervision to the annotation distribution (Sec.~\ref{sec:etf}), and control its data exposure through a 3-stage training schedule (Sec.~\ref{sec:schedule}).

\subsection{Co-DINO with a Swin-L backbone}
\label{sec:codino}

\begin{figure}[t!]
\centering
\includegraphics[width=0.90\linewidth]{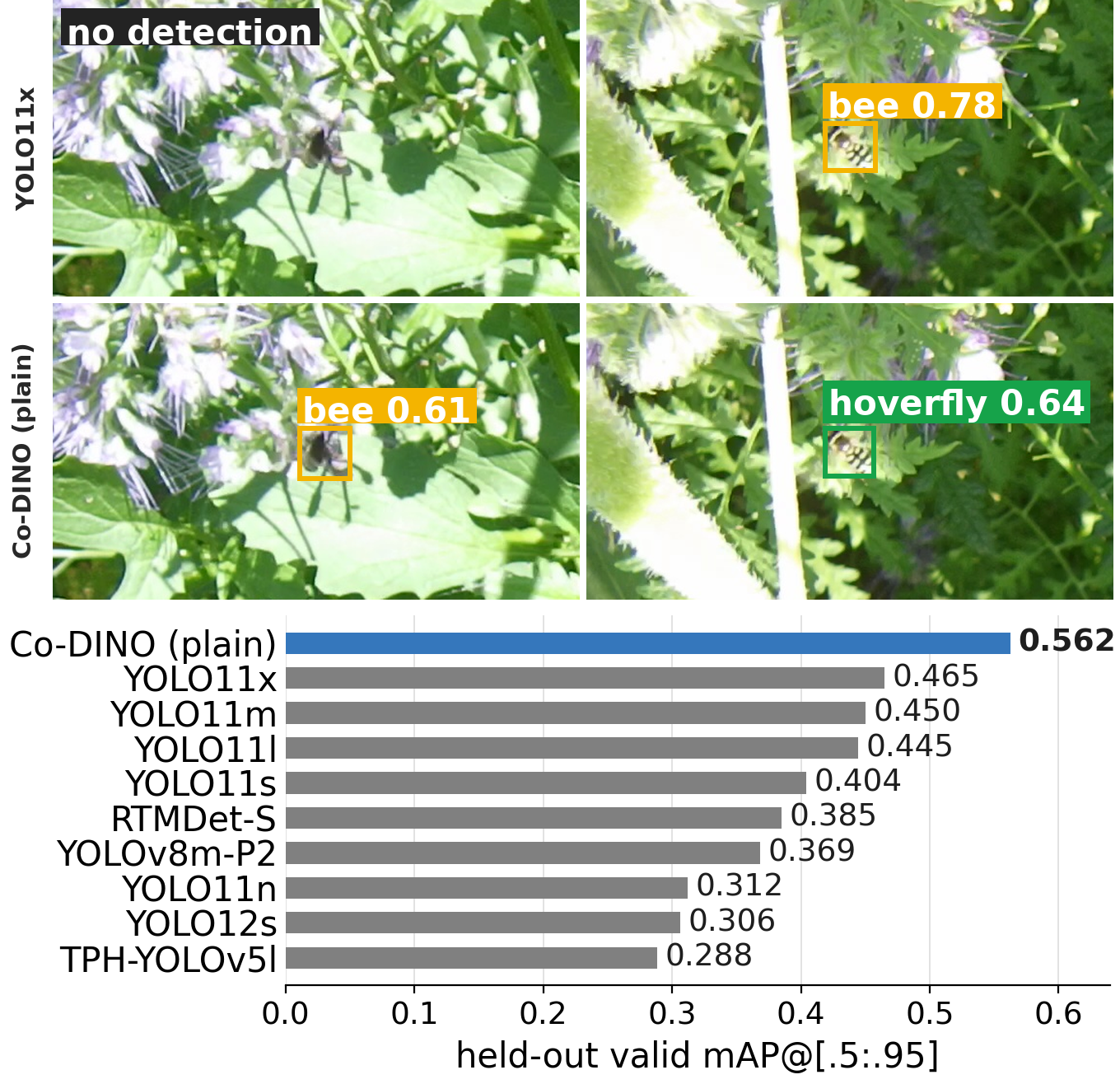}
\caption{\textbf{Detector-family comparison.} Top: YOLO11x misses a bee and confuses a hoverfly; yet, plain Co-DINO detects both at score $\geq0.5$. Bottom: recorded held-out mAP@$[.5{:}.95]$.}
\label{fig:teaser}
\end{figure}

On the 932 held-out keyframes of Fig.~\ref{fig:teaser}, plain Co-DINO with Swin-L reaches $0.562$ mAP against $0.465$ for the best of 9 YOLO and RTMDet configurations and corrects the shown miss and class confusion, so we select it~\cite{Zong2023CoDETR,Liu2021Swin}.

We keep Swin-L because its hierarchy retains a high-resolution output and multi-scale features. Swin-L produces 4 stage outputs, which the neck maps to 5 detector levels with 256 channels each. We initialize from an Objects365-to-COCO checkpoint~\cite{Shao2019Objects365,Lin2014COCO}.

Co-DINO extends DINO~\cite{Zhang2022DINO} with 2 auxiliary heads that run only during training: an RoI head and an ATSS head, each assigning many positives per object on the shared encoder features, which densifies supervision and forms extra training queries. The third category head is DINO's query head, which matches one query to one object through Hungarian assignment and alone serves inference; the auxiliary heads strengthen the shared features that the query head uses. All 3 category heads predict the same 4 classes, so the bee-dominated labels affect each training path.

\subsection{Fixed-ETF query regularization}
\label{sec:etf}

We apply fixed-ETF regularization only to the Co-DINO query head's Hungarian-matched final-layer queries. We retain its classifier and define 4 fixed, unit-norm simplex prototypes $\{\mathbf{m}_c\}_{c=1}^{4}\subset\mathbb{R}^{256}$ with equal pairwise angles~\cite{Papyan2020NeuralCollapse,Yang2022InducingNC,Yang2023NCFSCIL,Kim2025BeyondSyntheticReplays}. These equally separate the class targets; class weights change only their loss contribution.

\begin{figure}[t!]
\centering
\begin{tikzpicture}[
  font=\footnotesize,
  box/.style={draw, rounded corners=2pt, align=center, minimum height=8mm,
              text width=32mm, inner sep=2pt, fill=black!3},
  etfbox/.style={draw=cvprblue!85!black, rounded corners=2pt,
                 densely dashed, align=center, minimum height=8mm,
                 text width=32mm, inner sep=2pt, fill=cvprblue!9},
  etfloss/.style={etfbox, thick, fill=cvprblue!14},
  arr/.style={-{Latex[length=2mm]}, thick, black!70},
  darr/.style={-{Latex[length=2mm]}, thick, densely dashed,
               draw=cvprblue!85!black},
  lbl/.style={font=\scriptsize, text=black!65, inner sep=1.5pt}
]
\node[box] (queries) {last-layer decoder\\query states\\[-1pt]
  \scriptsize denoising queries excluded};
\node[box, below=5mm of queries] (heads) {learned classification\\+ box heads\\[-1pt]
  \scriptsize kept as the inference path};
\node[etfbox, right=6mm of heads] (match) {Hungarian assignment\\of predictions
  with GT\\yields indices and classes};
\node[etfbox, below=4mm of match] (proj) {selected query states\\linear projection\\
  + $\ell_2$ normalization};
\node[etfloss, below=4mm of proj] (loss) {class-weighted\\dot-regression loss};
\node[etfbox, left=6mm of proj]
  (target) {fixed simplex-ETF target\\of the matched class\\[-1pt]
  \scriptsize equal pairwise angles};
\draw[arr] (queries) -- (heads);
\draw[darr] (heads) -- (match);
\draw[darr] (match) -- node[lbl, midway, right=2pt] {indices} (proj);
\draw[darr] (proj) -- (loss);
\draw[darr] (target.south) |- (loss.west);
\draw[darr] (queries.east) -- node[lbl, above, pos=0.35] {query states}
  ++(4.3,0) |- (proj.east);
\end{tikzpicture}
\caption{\textbf{Fixed-ETF branch.} We exclude denoising queries before matching the final-layer predictions to ground truth. Matched indices select decoder states, matched classes select fixed targets, and dashed lines mark the training-only path.}
\label{fig:etf}
\end{figure}
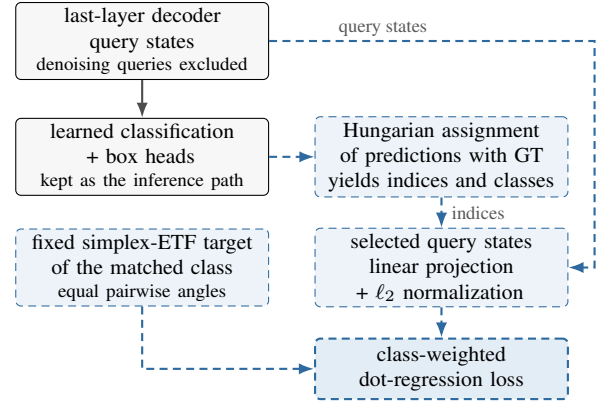

Following Fig.~\ref{fig:etf}, we exclude denoising queries, match final-layer predictions to ground truth, and use positive matches to select decoder states. We apply a learned bias-free linear projection, $\ell_2$-normalize its output as $\hat{\mathbf{p}}_i$, and compare it with the matched-class prototype $\mathbf{m}_{y_i}$. The class-weighted loss over matched set $\mathcal{P}$ is
\begin{equation}
\mathcal{L}_{\etf}=\frac{1}{\sum_{i\in\mathcal{P}}w_{y_i}}
\sum_{i\in\mathcal{P}}w_{y_i}\,
\tfrac{1}{2}\bigl(\langle\hat{\mathbf{p}}_i,
\mathbf{m}_{y_i}\rangle-1\bigr)^2 ,
\label{eq:etf}
\end{equation}
where $y_i$ is the class of match $i$ and $w_{y_i}$ is its class weight. The class weights are $w_c=1.0$, $1.5$, $4.0$, and $2.5$ for bee, bumblebee, hoverfly, and moth, weighting every rare class above bee. The total training loss then adds $0.1\,\mathcal{L}_{\etf}$ to Co-DINO's unchanged detection terms.

\subsection{3-stage final training}
\label{sec:schedule}

\begin{figure*}[t]
\centering
\begin{tikzpicture}[
  font=\footnotesize,
  stage/.style={draw, rounded corners=2pt, align=center, minimum height=12mm,
                minimum width=38mm, inner sep=3pt, fill=black!3},
  endpoint/.style={draw, rounded corners=2pt, align=center,
                   minimum height=9mm, minimum width=27mm,
                   inner sep=3pt, fill=cvprblue!10, thick},
  arr/.style={-{Latex[length=2mm]}, thick, black!70},
  note/.style={font=\scriptsize, align=center, text=cvprblue!80!black}
]
\node[stage] (replay) {All labeled keyframes\\\textbf{12 epochs}};
\node[stage, right=8mm of replay] (mosaic) {Class-aware crop mosaics\\
  \textbf{3 epochs}};
\node[stage, right=8mm of mosaic] (cool) {Original keyframes\\
  \textbf{2 epochs}\\[-1pt]\scriptsize low rate};
\node[endpoint, right=8mm of cool] (test) {Final model};
\draw[arr] (replay) -- (mosaic);
\draw[arr] (mosaic) -- (cool);
\draw[arr] (cool) -- (test);
\draw[thick, cvprblue!75!black]
  ($(replay.south west)+(0,-2mm)$) -- ($(cool.south east)+(0,-2mm)$);
\node[note, below=3.5mm of mosaic]
  {fixed-ETF loss active throughout all 3 training stages};
\end{tikzpicture}
\caption{\textbf{Training schedule.} We train for 12 epochs on the original keyframes, 3 with class-aware crop Mosaic, and 2 on the original keyframes at a lower learning rate.}
\label{fig:pipeline}
\end{figure*}
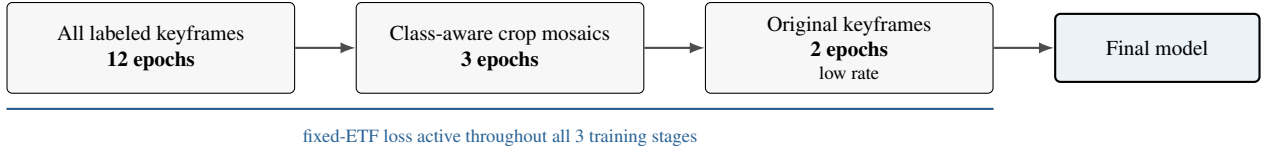

The loss above sets the supervision; what remains is the data each stage presents, which Fig.~\ref{fig:pipeline} organizes into 3 stages. We first adapt the pretrained detector to the BuzzSpot classes and full-frame context by training on all $6{,}207$ train+valid keyframes for 12 epochs at learning rate $10^{-4}$.

\begin{figure}[t!]
\centering
\includegraphics[width=0.98\linewidth]{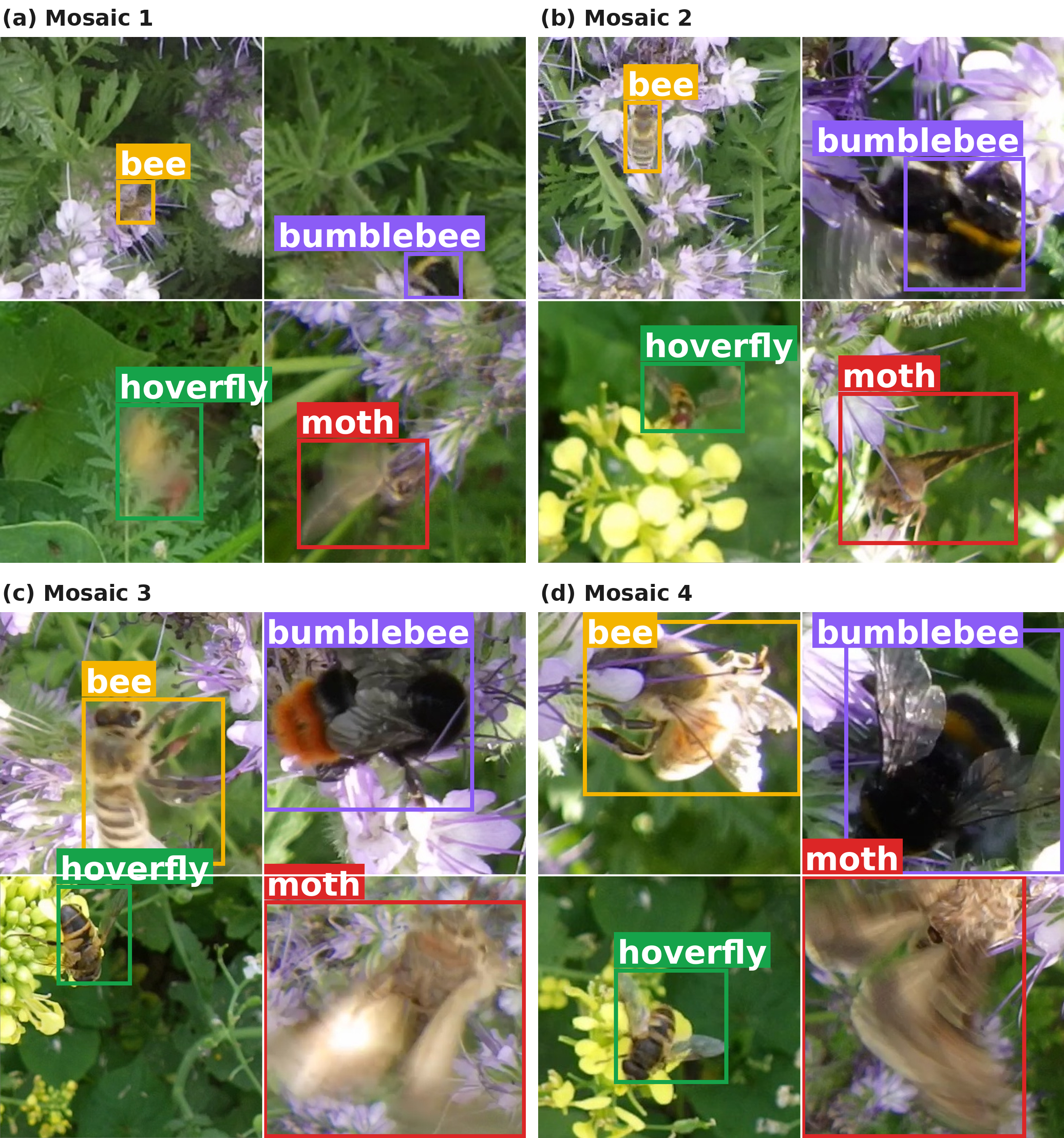}
\caption{\textbf{Visualization of mosaic augmentation.}}
\label{fig:mosaic_views}
\end{figure}

For stage 2, we train for 3 epochs at $10^{-5}$ on a class-aware crop-mosaic pool. We build this pool by sampling $5{,}993$ object-centered views from the original training split, with base class multipliers $[0.75,10,3,10]$ for bee, bumblebee, hoverfly, and moth and square crop side lengths of $192$--$384$ pixels. Heuristic factors then adjust each annotation's sampling weight: it doubles when the box area falls below that of a $32{\times}32$-pixel square, multiplies by $1.5$ for blur, and by another $1.5$ for occlusion level 2 or higher. The crop center jitters by up to $15\%$ of its side length along each axis, and every source box whose intersection retains at least $35\%$ of its original area is remapped with its category and attributes preserved. All 932 uncropped validation images also join the pool to retain full-frame samples. Tab.~\ref{tab:composition} reports its composition before Mosaic: the 3 rare classes account for $55.1\%$ of the annotations, compared with $19.9\%$ in the original data. Fig.~\ref{fig:mosaic_views} shows 4 views on a $2048{\times}2048$ Mosaic canvas with a $1024{\times}1024$ base scale~\cite{Bochkovskiy2020YOLOv4}.

\begin{figure}[t!]
\centering
\includegraphics[width=0.98\linewidth]{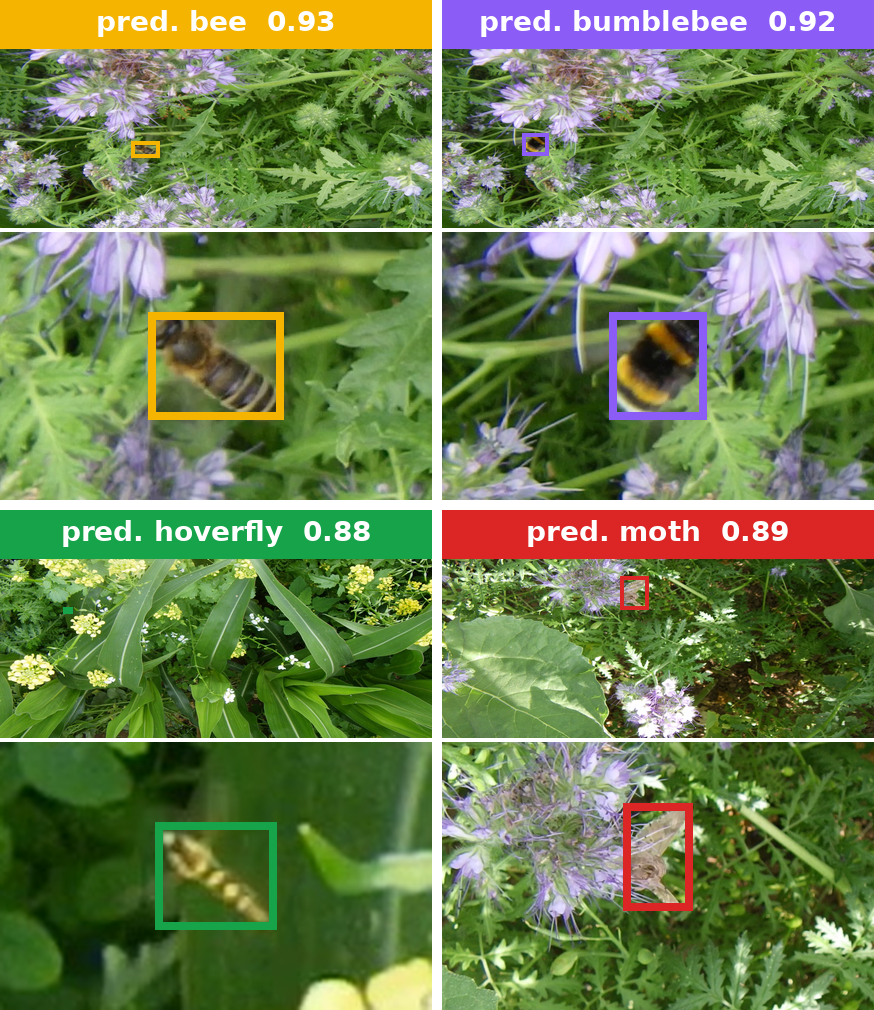}
\caption{\textbf{FinalTest predictions.}}
\label{fig:finaltest_outputs}
\end{figure}

Mosaic changes the full-frame context and scale distribution, so we finish with 2 epochs on the original train+valid keyframes at a lower learning rate of $5{\times}10^{-6}$.

\section{Experiments}
\label{sec:exp}

\subsection{Dataset and protocol}

\begin{table}[b!]
\centering
\small
\setlength{\tabcolsep}{2.8pt}
\begin{tabular}{@{}lrrrrr@{}}
\toprule
Training source & Bee & B.bee & Hover. & Moth & Rare share \\
\midrule
Original train+valid & 9,611 & 267 & 1,848 & 274 & 19.9\% \\
Crop-mosaic pool & 3,420 & 799 & 2,628 & 777 & 55.1\% \\
\bottomrule
\end{tabular}
\caption{\textbf{Class composition.} Counts precede Mosaic augmentation; rare share combines bumblebee, hoverfly, and moth.}
\label{tab:composition}
\end{table}

The train+valid annotations used for final training contain $6{,}207$ keyframes and $12{,}000$ boxes across the 4 BuzzSpot classes; FinalTest contains $4{,}763$ keyframes with hidden annotations. For scale analysis, we use the COCO API's small-object range and count boxes whose area is at most that of a $32{\times}32$-pixel square as small~\cite{COCOAPI}. This includes $1{,}018$ of $1{,}848$ hoverfly boxes ($55.1\%$), compared with 425 of $9{,}611$ bee boxes ($4.4\%$), 1 of 267 bumblebee boxes ($0.4\%$), and 0 of 274 moth boxes. For detector selection, we train each model on the training videos and evaluate it on the 932 validation keyframes. We average COCO-style mAP over IoU thresholds from $0.50$ to $0.95$ in $0.05$ steps. We then train the final model on all train+valid keyframes. With the validation split now in-sample, FinalTest remains the only held-out evaluation.

\subsection{Implementation details}

Our released recipe trains data-parallel with PyTorch~\cite{Li2020PyTorchDistributed} on 4 NVIDIA A100 80 GB GPUs at batch size 1 per GPU. We optimize with AdamW, use weight decay $10^{-4}$, clip the gradient norm at $0.1$, and set the backbone learning rate to $0.1$ times each stage's rate from Sec.~\ref{sec:schedule}. Every stage follows iteration-wise cosine decay, with linear warm-up over the first 250 iterations in stage 1 and 100 in stages 2 and 3. The query head weights its quality-focal classification~\cite{Li2020GFL}, L1 box, and generalized IoU~\cite{Rezatofighi2019GIoU} terms by $1$, $5$, and $2$. In stages 1 and 3, we apply horizontal flipping with probability $0.5$ and Co-DINO's 2-branch multi-scale resize/crop augmentation. In stage 2, we replace this pipeline with Mosaic at probability $0.85$, photometric distortion, the same horizontal flipping, and multi-scale resizing. At inference, we resize each keyframe, preserving its aspect ratio, to fit within $2048{\times}1280$, and run one full-frame pass.

\subsection{Detector-family comparison}
\label{sec:family}

Fig.~\ref{fig:teaser} reports the recorded held-out mAP for plain Co-DINO and 9 configurations spanning TPH-YOLOv5~\cite{Zhu2021TPHYOLOv5}, YOLOv8~\cite{Jocher2023YOLOv8}, RTMDet~\cite{Lyu2022RTMDet}, YOLO11~\cite{Jocher2024YOLO11}, and YOLOv12~\cite{Tian2025YOLOv12}. Because the configurations use different protocols, we use the comparison only for detector selection.

For plain Co-DINO in Fig.~\ref{fig:teaser}, class AP@$[.5{:}.95]$ is $[0.616, 0.714, 0.328, 0.592]$ for bee, bumblebee, hoverfly, and moth: hoverfly trails every other class and carries the highest small-box share in the annotations. This weakness motivates stage 2's emphasis on rare and small objects.

\subsection{FinalTest result}

Our final checkpoint reaches $0.5062$ mAP@$[.5{:}.95]$ on FinalTest, $0.0197$ mAP ($4.1\%$ relative) above the second entry. Fig.~\ref{fig:finaltest_outputs} shows selected FinalTest outputs.

\section{Conclusion}
\label{sec:conclusion}

We address the challenge's 2 difficulties with a single detector. To handle the few-pixel insects, we select Co-DINO with Swin-L from 10 configurations; to handle the bee-dominated labels, we fine-tune on class-aware crop mosaics and add fixed-ETF supervision. For practical use, we avoid ensembling and test-time augmentation, and take rank 1 on FinalTest at $0.5062$ mAP@$[.5{:}.95]$.

\noindent\textbf{Additional explorations.} We also test class-balanced sampling, object-crop copy-paste fine-tuning~\cite{Ghiasi2021SimpleCopyPaste}, detector pseudo-labels~\cite{Kim2024VLMPL}, and synthetic image augmentation adapted from prior work~\cite{Kim2025SyntheticImages}. All underperform the baseline, so we exclude them from the final recipe.

{\small\noindent\textbf{Acknowledgement.} This work is supported by IITP grants (No. RS2020-II201336 Artificial intelligence graduate school program (UNIST) 34\%; No. RS-2022-II220264 Comprehensive video understanding and generation with knowledge-based deep logic neural network 33\%, No. RS-2025-25442149 LG AI STAR Talent Development Program for Leading Large-Scale Generative AI Models in the Physical AI Domain 33\%), funded by the Korean government (MSIT).\par}

\FloatBarrier

{\small
\bibliographystyle{unsrtnat}
\bibliography{references}

\begin{thebibliography}{24}
\providecommand{\natexlab}[1]{#1}
\providecommand{\url}[1]{\texttt{#1}}
\expandafter\ifx\csname urlstyle\endcsname\relax
  \providecommand{\doi}[1]{doi: #1}\else
  \providecommand{\doi}{doi: \begingroup \urlstyle{rm}\Url}\fi

\bibitem[{CVPPA@ECCV}(2026)]{BuzzSpot2026Challenge}
{CVPPA@ECCV}.
\newblock {BuzzSpot} challenge, 2026.

\bibitem[Zong et~al.(2023)Zong, Song, and Liu]{Zong2023CoDETR}
Zhuofan Zong, Guanglu Song, and Yu~Liu.
\newblock {DETR}s with collaborative hybrid assignments training.
\newblock In \emph{ICCV}, 2023.

\bibitem[Liu et~al.(2021)Liu, Lin, Cao, Hu, Wei, Zhang, Lin, and
  Guo]{Liu2021Swin}
Ze~Liu, Yutong Lin, Yue Cao, Han Hu, Yixuan Wei, Zheng Zhang, Stephen Lin, and
  Baining Guo.
\newblock Swin transformer: Hierarchical vision transformer using shifted
  windows.
\newblock In \emph{ICCV}, 2021.

\bibitem[Kim et~al.(2024{\natexlab{a}})Kim, Hong, Kim, Kim, Tiruneh, On, Song,
  Choi, and Baek]{Kim2024ClassWiseBuffer}
Junsu Kim, Sumin Hong, Chanwoo Kim, Jihyeon Kim, Yihalem~Yimolal Tiruneh,
  Jeongwan On, Jihyun Song, Sunhwa Choi, and Seungryul Baek.
\newblock Class-wise buffer management for incremental object detection: An
  effective buffer training strategy.
\newblock In \emph{ICASSP}, 2024{\natexlab{a}}.

\bibitem[Bochkovskiy et~al.(2020)Bochkovskiy, Wang, and
  Liao]{Bochkovskiy2020YOLOv4}
Alexey Bochkovskiy, Chien-Yao Wang, and Hong-Yuan~Mark Liao.
\newblock {YOLOv4}: Optimal speed and accuracy of object detection.
\newblock \emph{arXiv}, 2020.

\bibitem[Yang et~al.(2022)Yang, Chen, Li, Xie, Lin, and
  Tao]{Yang2022InducingNC}
Yibo Yang, Shixiang Chen, Xiangtai Li, Liang Xie, Zhouchen Lin, and Dacheng
  Tao.
\newblock Inducing neural collapse in imbalanced learning: Do we really need a
  learnable classifier at the end of deep neural network?
\newblock In \emph{NeurIPS}, 2022.

\bibitem[Yang et~al.(2023)Yang, Yuan, Li, Lin, Torr, and Tao]{Yang2023NCFSCIL}
Yibo Yang, Haobo Yuan, Xiangtai Li, Zhouchen Lin, Philip H.~S. Torr, and
  Dacheng Tao.
\newblock Neural collapse inspired feature-classifier alignment for few-shot
  class-incremental learning.
\newblock In \emph{ICLR}, 2023.

\bibitem[Kim et~al.(2025{\natexlab{a}})Kim, Ku, Han, and
  Baek]{Kim2025BeyondSyntheticReplays}
Junsu Kim, Yunhoe Ku, Dongyoon Han, and Seungryul Baek.
\newblock Beyond synthetic replays: Turning diffusion features into few-shot
  class-incremental learning knowledge.
\newblock \emph{arXiv}, 2025{\natexlab{a}}.

\bibitem[Shao et~al.(2019)Shao, Li, Zhang, Peng, Yu, Zhang, Li, and
  Sun]{Shao2019Objects365}
Shuai Shao, Zeming Li, Tianyuan Zhang, Chao Peng, Gang Yu, Xiangyu Zhang, Jing
  Li, and Jian Sun.
\newblock {Objects365}: A large-scale, high-quality dataset for object
  detection.
\newblock In \emph{ICCV}, 2019.

\bibitem[Lin et~al.(2014)Lin, Maire, Belongie, Hays, Perona, Ramanan,
  Doll{\'a}r, and Zitnick]{Lin2014COCO}
Tsung-Yi Lin, Michael Maire, Serge Belongie, James Hays, Pietro Perona, Deva
  Ramanan, Piotr Doll{\'a}r, and C.~Lawrence Zitnick.
\newblock Microsoft {COCO}: Common objects in context.
\newblock In \emph{ECCV}, 2014.

\bibitem[Zhang et~al.(2023)Zhang, Li, Liu, Zhang, Su, Zhu, Ni, and
  Shum]{Zhang2022DINO}
Hao Zhang, Feng Li, Shilong Liu, Lei Zhang, Hang Su, Jun Zhu, Lionel~M. Ni, and
  Heung-Yeung Shum.
\newblock {DINO}: {DETR} with improved denoising anchor boxes for end-to-end
  object detection.
\newblock In \emph{ICLR}, 2023.

\bibitem[Papyan et~al.(2020)Papyan, Han, and Donoho]{Papyan2020NeuralCollapse}
Vardan Papyan, X.~Y. Han, and David~L. Donoho.
\newblock Prevalence of neural collapse during the terminal phase of deep
  learning training.
\newblock \emph{PNAS}, 2020.

\bibitem[{COCO API}(2015)]{COCOAPI}
{COCO API}.
\newblock {Microsoft COCO Evaluation Software}, 2015.

\bibitem[Li et~al.(2020{\natexlab{a}})Li, Zhao, Varma, Salpekar, Noordhuis, Li,
  Paszke, Smith, Vaughan, Damania, and Chintala]{Li2020PyTorchDistributed}
Shen Li, Yanli Zhao, Rohan Varma, Omkar Salpekar, Pieter Noordhuis, Teng Li,
  Adam Paszke, Jeff Smith, Brian Vaughan, Pritam Damania, and Soumith Chintala.
\newblock {PyTorch Distributed}: Experiences on accelerating data parallel
  training.
\newblock \emph{PVLDB}, 2020{\natexlab{a}}.

\bibitem[Li et~al.(2020{\natexlab{b}})Li, Wang, Wu, Chen, Hu, Li, Tang, and
  Yang]{Li2020GFL}
Xiang Li, Wenhai Wang, Lijun Wu, Shuo Chen, Xiaolin Hu, Jun Li, Jinhui Tang,
  and Jian Yang.
\newblock Generalized focal loss: Learning qualified and distributed bounding
  boxes for dense object detection.
\newblock In \emph{NeurIPS}, 2020{\natexlab{b}}.

\bibitem[Rezatofighi et~al.(2019)Rezatofighi, Tsoi, Gwak, Sadeghian, Reid, and
  Savarese]{Rezatofighi2019GIoU}
Hamid Rezatofighi, Nathan Tsoi, JunYoung Gwak, Amir Sadeghian, Ian Reid, and
  Silvio Savarese.
\newblock Generalized intersection over union: A metric and a loss for bounding
  box regression.
\newblock In \emph{CVPR}, 2019.

\bibitem[Zhu et~al.(2021)Zhu, Lyu, Wang, and Zhao]{Zhu2021TPHYOLOv5}
Xingkui Zhu, Shuchang Lyu, Xu~Wang, and Qi~Zhao.
\newblock {TPH-YOLOv5}: Improved {YOLOv5} based on transformer prediction head
  for object detection on drone-captured scenarios.
\newblock In \emph{ICCVW}, 2021.

\bibitem[{Ultralytics}(2023)]{Jocher2023YOLOv8}
{Ultralytics}.
\newblock {YOLOv8} software, 2023.

\bibitem[Lyu et~al.(2022)Lyu, Zhang, Huang, Zhou, Wang, Liu, Zhang, and
  Chen]{Lyu2022RTMDet}
Chengqi Lyu, Wenwei Zhang, Haian Huang, Yue Zhou, Yudong Wang, Yanyi Liu,
  Shilong Zhang, and Kai Chen.
\newblock {RTMDet}: An empirical study of designing real-time object detectors.
\newblock \emph{arXiv}, 2022.

\bibitem[{Ultralytics}(2024)]{Jocher2024YOLO11}
{Ultralytics}.
\newblock {YOLO11} software, 2024.

\bibitem[Tian et~al.(2025)Tian, Ye, and Doermann]{Tian2025YOLOv12}
Yunjie Tian, Qixiang Ye, and David Doermann.
\newblock {YOLOv12}: Attention-centric real-time object detectors.
\newblock In \emph{NeurIPS}, 2025.

\bibitem[Ghiasi et~al.(2021)Ghiasi, Cui, Srinivas, Qian, Lin, Cubuk, Le, and
  Zoph]{Ghiasi2021SimpleCopyPaste}
Golnaz Ghiasi, Yin Cui, Aravind Srinivas, Rui Qian, Tsung-Yi Lin, Ekin~D.
  Cubuk, Quoc~V. Le, and Barret Zoph.
\newblock Simple copy-paste is a strong data augmentation method for instance
  segmentation.
\newblock In \emph{CVPR}, 2021.

\bibitem[Kim et~al.(2024{\natexlab{b}})Kim, Ku, Kim, Cha, and
  Baek]{Kim2024VLMPL}
Junsu Kim, Yunhoe Ku, Jihyeon Kim, Junuk Cha, and Seungryul Baek.
\newblock {VLM-PL}: Advanced pseudo labeling approach for class incremental
  object detection via vision-language model.
\newblock In \emph{CVPRW}, 2024{\natexlab{b}}.

\bibitem[Kim et~al.(2025{\natexlab{b}})Kim, Ku, and
  Baek]{Kim2025SyntheticImages}
Junsu Kim, Yunhoe Ku, and Seungryul Baek.
\newblock Can synthetic images conquer forgetting? {Beyond} unexplored doubts
  in few-shot class-incremental learning.
\newblock In \emph{ICCVW}, 2025{\natexlab{b}}.

\end{thebibliography}
}

\end{document}